\documentclass[aoas]{imsart}

\RequirePackage{amsthm,amsmath,amsfonts,amssymb}
\RequirePackage[authoryear]{natbib}
\RequirePackage[colorlinks,citecolor=blue,urlcolor=blue]{hyperref}
\RequirePackage{graphicx}
\RequirePackage{booktabs}

\usepackage{booktabs}       
\usepackage{nicefrac}       
\usepackage{microtype}      
\usepackage[ruled,vlined]{algorithm2e}
\usepackage{multirow}
\usepackage{float}
\usepackage{bm}
\usepackage{mathrsfs}
\usepackage{subcaption}
\usepackage{comment}
\usepackage{tabularx}
\usepackage[export]{adjustbox}
\usepackage[table]{xcolor} 

\startlocaldefs

\newcommand{\sumij}{\sum\limits_{i=1}^{I}\sum\limits_{j=1}^{J}}

\newcommand{\sumi}{\sum\limits_{i=1}^{I}}
\newcommand{\sumj}{\sum\limits_{j=1}^{J}}

\newcommand{\prodi}{\prod\limits_{i=1}^{I}}
\newcommand{\prodj}{\prod\limits_{j=1}^{J}}
\newcommand{\prodq}{\prod\limits_{q=1}^{Q}}

\endlocaldefs

\begin{document}

\begin{frontmatter}
\title{Variational Inference for Additive Main  and Multiplicative Interaction Effects Models}
\runtitle{Variational Inference for Additive Main  and Multiplicative Interaction Effects Models}

\begin{aug}
\author[A]{\fnms{Antônia A. L.} \snm{dos Santos}\ead[label=e1]{ antonia.lemosdossantos.2020@mumail.ie }},
\author[A]{\fnms{Rafael A.} \snm{Moral}\ead[label=e1]{ rafael.deandrademoral@mu.ie}},
\author[A]{\fnms{Danilo A.} \snm{Sarti}\ead[label=e1]{daniloasarti@gmail.com}},
\author[A, B]{\fnms{Andrew C.} \snm{Parnell}\ead[label=e1]{Andrew.parnell@mu.ie}}
\address[A]{Hamilton Institute, Department of Mathematics and Statistics, Maynooth University, Ireland}
\address[B]{Insight Centre for Data Analytics, Maynooth University, Ireland}

\end{aug}

\begin{abstract}
	In plant breeding the presence of a genotype by environment (\verb+GxE+) interaction has a strong impact on cultivation decision making and the introduction of new crop cultivars. The combination of linear and bilinear terms has been shown to be very useful in modelling this type of data. A widely-used approach to identify \verb+GxE+ is the Additive Main Effects and Multiplicative Interaction Effects (AMMI) model. However, as data frequently can be high-dimensional, Markov chain Monte Carlo (MCMC) approaches can be computationally infeasible. In this article, we consider a variational inference approach for such a model. We derive variational approximations for estimating the parameters and we compare the approximations to MCMC using both simulated and real data. The new inferential framework we propose is on average two times faster whilst maintaining the same predictive performance as MCMC.
\end{abstract}

\begin{keyword}
\kwd{AMMI models \and Genotype x enviroment interaction}
\kwd{Variational inference}
\kwd{Bayesian inference}
\end{keyword}

\end{frontmatter}

\section{Introduction}

In plant breeding, it is often of interest to identify which genotypes perform best in different environments. The presence of genotype $\times$ environment (\verb+GxE+) interactions is an important factor, and has a strong impact on the yield. Furthermore, it contributes to the improvement of breeding programs \citep{mclaren1994use}. Many authors deal with modelling interactions in plant genomic data. In particular, \cite{crossa2010linear} review various statistical models for analyzing \verb+GxE+ interactions. A common approach is to create models that combine linear and bilinear (interaction) terms. Often, these bilinear interaction terms are not simple multiplicative combinations of genotype-environment pairs, but rather latent parameters to be estimated. For papers on linear-bilinear models, see \cite{gauch2008statistical}, \cite{crossa2010linear}, \cite{poland2012genomic} and \cite{gauch2013simple}. 

One of the most widely used linear-bilinear models is the Additive Main  and Multiplicative Interaction (AMMI) effects model \citep{gauch1988model, gauch1992statistical}. AMMI incorporates both additive and multiplicative  components from the two-way data structure, first educing the principal additive components, and then investigating the \verb+GxE+ component with principal components analysis. The results of an AMMI-style analysis are usually displayed graphically in biplots \citep{gauch1997identifying, yan2000cultivar, yan2002biplot}, that help to interpret the \verb+GxE+ interactions. Further, these models typically perform well with regard to their predictive properties \citep{gauch2006statistical, gauch2008statistical}. 

Several approaches have been made in estimating the parameters of the AMMI model, mostly from the point of view of classical inference \citep{gilbert1963non, gabriel1978least, gauch1988model, van1995linear}. However, the desirable characteristics for evaluating uncertainty and including expert knowledge, as a priori information, led the Bayesian approach to be applied in \verb+GxE+ modelling \citep{foucteau2001statistical, theobald2002bayesian, cotes2006bayesian}, and consequently, in the AMMI model. In the latter case, it is necessary to deal with the constraints associated with the model, and the problems caused by the orthonormal bases used in the decomposition of the bilinear term. Due to these restrictions, complicated prior distribution decisions for AMMI inference may be necessary \citep{viele2000parsimonious, crossa2011bayesian}. An alternative approach proposed by \cite{josse2014another} ignores these restrictions on the prior distributions and applies later-level processing of the posterior.

When the data sets are particularly large or when models become complex, traditional methods for Bayesian inference can fail to sample from the posterior distribution. An alternative way to deal with these issues is to use methods such as Variational Inference (VI); also called Variational Bayes \citep{beal2003variational, ormerod2010explaining, blei2017variational}. This approach has been widely used in recent years because, whilst the variational approximations do not always converge to the exact posterior distribution, they are computationally much faster. In some agricultural models, for example, VI has been demonstrated to perform well \citep{montesinos2017variational,gillberg2019modelling}. 

To date we have not found a paper that applies VI to AMMI models, and this is the focus of our paper. We follow \cite{josse2014another}'s approach for constructing the AMMI model, but fit the model using  VI in order to enjoy the characteristics of the Bayesian approach and reduce the computational cost in the approximation of the posteriors of the model. In a set of simulation studies, we find that VI performs well in terms of speed and, in comparison with the results obtained via Markov chain Monte Carlo (MCMC), our results were similar in terms of accuracy. Unsurprisingly we find the speed boost to be greater when the data set grows larger. 

Our paper is structured as follows. We begin in Section \ref{ammi} by reviewing the AMMI model, which is formulated in the Bayesian style. Section \ref{vb} provides a review of VI. In Section \ref{vammi} we present the mathematics of the variational updates for the AMMI model. In Section \ref{results} we evaluate our methods and make comparisons with MCMC using simulated and real data. We present graphs comparing the results via the two methods, as well as the prediction results. In Section \ref{discussion} we provide a discussion of our findings and point to future research. An R implementation of the AMMI models and variational inference of the model used to run all experiments and generate the results of the paper is  available on Github here (\url{https://github.com/Alessandra23/vammi}).

\section{Theoretical Background}

\subsection{AMMI Model}\label{ammi}

In this section, we define the AMMI model and  present the restrictions that guarantee identifiability of the model parameters.  The AMMI model for an outcome variable (e.g. log yield per hectare), $y_{ij} \in \mathbb{R}$, is formulated as
\begin{eqnarray}\label{eq:yij}
y_{ij} &=& \mu +g_{i}+e_{j}+\sum_{q=1}^{Q} \lambda_{q} \gamma_{iq} \delta_{jq}+\epsilon_{ij},
\end{eqnarray}

\noindent
where $\mu$ is the grand mean, $g_i$, $i \in \{1, \dots, I\}$ and $e_j$, $j \in \{1, \dots , J\}$, represent the effect of the $i$-th genotype and $j$-th environment, respectively, and $\epsilon_{ij}$ is noise distributed as $\mathcal{N}(0,\sigma^2)$. The term in the summation is called the bilinear component and captures the interaction effects, where $\lambda_q$ is the singular value of the $q$-th bilinear component, and $\gamma_{iq}$ and $\delta_{jq}$ are the left and right singular vectors of the interaction, respectively. The upper term of the summation $Q$ represents the number of bilinear terms to include and is fixed before running the model. For visualization and interpretability issues it is common to set $Q$ less than 3. All the terms on the right hand side of the equation (with the exception of $Q$) are parameters to be estimated. The model can be trivially extended to account for block or replicate effects, but we do not explore such extensions in this paper. 

In matrix form, denoting $\bm{Y} = (y_{ij}) \in \mathbb{R}^{I\times J}$, Equation (\ref{eq:yij}) is equivalent to
\begin{eqnarray}\label{eq:yijmatrix}
\bm{Y} &=& \mu\bm{1}_I\bm{1}'_J + \bm{g}\bm{1}'_J + \bm{1}_I \bm{e}' + \bm{\Gamma}\bm{\Lambda}\bm{\Delta}' +\bm{E}.
\end{eqnarray}
where $\bm{\Gamma} \in \mathbb{R}^{I\times Q}, \bm{\Delta}\in \mathbb{R}^{J\times Q}$, $\bm{\Lambda}\in \mathbb{R}^{Q\times Q}$,  $\bm{1}_m$ is a column vector of ones of size $m$. $\bm{\gamma}_q$ and $\bm{\delta}_q$ arise from the $q$-th column of matrices $\bm{\Gamma}$ and $\bm{\Delta}$ respectively. $\boldsymbol{\Lambda}$ is a diagonal matrix consisting of the terms $\lambda_1, \ldots, \lambda_Q$. The other bold terms indicate vector stacking of the individual main effects.

The nature of the over-parameterization means that it is necessary to establish some conditions of identifiability and interpretability for the parameters. It is thus common to assume the constraints below:
\begin{enumerate}
    \item[(i)] $\bm{1}_I'\bm{g} = \bm{1}_J'\bm{e} = 0$,
    \item[(ii)] $\bm{1}_I'\bm{\gamma}_q = \bm{1}_J'\bm{\delta}_q = 0$ for all $q \in \{1, \dots, Q\}$,
    \item[(iii)] $\bm{\Gamma'\Gamma} = \bm{\Delta'\Delta} = \bm{\mathrm{I}}_Q$,
    \item[(iv)] the diagonal terms of $\bm{\Lambda}$ are ordered such that $\lambda_{1} \geq \lambda_{2} \geq \dots \geq \lambda_{Q} \geq 0$,
    \item[(v)] the first entry of each column of $\bm{\Gamma}$ is positive.
\end{enumerate}
The constraints (i) and (ii) are interpretability constraints, whereas (iii), (iv) and (v) are necessary constraints to ensure identifiability of the model.

From a frequentist perspective, a multi-stage least squares method can be used to estimate the parameters of the model, first estimating the linear terms $\mu +g_{i}+e_{j}$ and then using Principal Components Analysis (PCA), equivalent to maximum likelihood estimation, to estimate the residuals of the nonlinear term and so create $\sum_{q=1}^{Q} \lambda_{q} \gamma_{iq} \delta_{jq}$ \citep{gilbert1963non, gollob1968statistical, gabriel1978least}.

From a Bayesian perspective, the restrictions can be dealt with via the prior distributions. In \cite{cornelius1999prediction} a Bayesian shrinkage estimator is proposed. \cite{crossa2011bayesian} introduce proper prior distributions and use a Gibbs sampler for inference on the parameters to target the exact posterior distribution. \cite{perez2012general} propose the von Mises–Fisher distribution as a prior for the orthonormal  matrices $\bm{\Gamma}$ and $\bm{\Delta}$. By contrast, \cite{josse2014another} propose that the over-parameterization be handled subsequent to the model fitting, initially ignoring the problems at the prior level and applying an appropriate post-processing of the posterior distribution. They argue that this approach allows easily interpretable inferences to be obtained, in addition to simpler implementation in statistical software. \cite{mendescomparing} and \cite{omer2017comparing} perform a comparison between the classical and Bayesian approaches of the AMMI model, showing that the Bayesian modelling is more flexible than the classical one. More recently, \cite{sarti2021bayesian} propose the use of semi-parametric Bayesian Additive Regression Trees (BART) to capture  genotypic by environment interactions.

One of the main problems when using the Bayesian methodology in linear-bilinear models is the associated computational cost, given the complex structure of the parameters. Taking this into account, we follow the proposal of \cite{josse2014another}, applying the prior distributions suggested in the matrices of the linear term, and using the variational inference to obtain the estimates of the AMMI model parameters, thus reducing the computational time of the fitting process.

\subsection{Variational Inference}\label{vb}

The basic idea of VI, as an alternative to Markov chain Monte Carlo, is to approximate the posterior distribution via optimization, thereby making the estimation process computationally faster. First, we choose a family of approximate densities over a set of latent variables which we believe will provide a good approximation to the true posterior. Then we find the set of parameters that make our approximation as close as possible to the posterior distribution. \cite{blei2017variational} reviews the method, presenting examples, applications, and a discussion of problems and current research on the topic. 

Let $\bm{y} = y_1, ..., y_n$ a set of observations, $\bm{\theta}$ a vector of parameters, $p(y)$ the marginal distribution of observations, and $p(\bm{y}, \bm{\theta})$ the joint density of the model and the parameters. The density transform approach, one of the most common variants of VI \citep{ormerod2010explaining}, consists of approximating the posterior distribution of $p(\bm{\theta}|\bm{y}) = p(\bm{y}, \bm{\theta})/p(\bm{y})$ by a distribution $q(\bm{\theta})$ from a set of tractable distributions $\mathscr{Q}$, for which we minimize the Kullback-Leibler (KL) divergence. Since it is not possible to calculate $\mathrm{KL}$ directly, the optimization is done over an equivalent quantity, called the evidence lower bound (ELBO), where maximizing it is equivalent to minimizing the $\mathrm{KL}$ divergence:
\begin{eqnarray}
\operatorname{ELBO}[q(\boldsymbol{\theta})] &=& -\mathrm{KL}[q(\boldsymbol{\theta}) \| p(\boldsymbol{\theta} \mid \mathbf{y})]+\log p(\mathbf{y}).
\end{eqnarray}

A broad class of distributions to describe the variational family $\mathscr{Q}$ is the mean-field approximation, which assumes that the elements of $\bm{\theta}$ are mutually independent, then $q(\bm{\theta})$ can be factored into $\prod_{m=1}^M q_m(\bm{\theta}_m)$, where $M$ is the number of partitions of the vector of parameters $\bm{\theta}$. Therefore, each variable $\bm{\theta}_m$ can be governed by its own variational factor. Bringing together the ELBO and the mean-field family, the optimal $q_m(\bm{\theta}_m)$ are proportional to 
\begin{eqnarray}
q_m(\bm{\theta}_m) &\propto& \exp\{ E_{-\bm{\theta}_m} \log p(\bm{\theta}_m|\bm{y}, \bm{\theta}_{-\bm{\theta}_m})\},
\end{eqnarray}
\noindent
where $E_{-\bm{\theta}_m}$ denotes the expectation with respect to all $\bm{\theta}$ variational distributions except $\bm{\theta}_m$. If the prior distributions are conjugate it is possible to obtain explicit expressions for each component leading to simple parameter updates, which is one of the advantages of the approach. As a disadvantage, \cite{montesinos2017variational} comments that the restrictions imposed by the mean-field approximation can lead to underestimation of the variability of parameter estimates,  and further emphasizes that it is not a good option if there is strong dependency between the parameters.

\subsection{VI applied to the AMMI Model}\label{vammi}

Exact inference for the model is intractable and we use 
the variational approximation for computational efficiency. Following \cite{josse2014another}, we list the priors defined for the complete set of parameters without considering any hard constrains (Table \ref{table:priorVar}). To simplify notation, let $\Theta = \{\mu,\bm{g},\bm{e},\bm{\lambda}, \bm{\gamma},\bm{\delta},\bm{\sigma^2}\}$ be the set of parameters and let $\bm{\theta} = \{\mu_\mu, \sigma^2_\mu, \sigma^2_g, \sigma^2_e, \sigma^2_\lambda, a, b\}$ be all the hyper-parameters.  Omitting the dependency on $\bm{\theta}$, we state the joint posterior density of the parameter vector of the variational AMMI model as
\begin{eqnarray*}
p(\bm{Y}, \mu , \bm{g}, \bm{e}, \bm{\Lambda}, \bm{\Gamma}, \bm{\Delta}, \sigma^2) &\propto& p(\bm{Y}| \mu, \bm{g}, \bm{e}, \bm{\Lambda}, \bm{\Gamma}, \bm{\Delta}, \sigma^2) p(\mu) p(\bm{g}) p(\bm{e}) p(\bm{\Lambda}) p(\bm{\Gamma}) p(\bm{\Delta}) p(\sigma^2),
\end{eqnarray*}
and assume that the posterior distribution is approximated by factorizing the variational approximation,
\begin{eqnarray*}
p(\mu , \bm{g}, \bm{e}, \bm{\Lambda}, \bm{\Gamma}, \bm{\Delta}, \sigma^2) &\propto& q(\mu) q(\bm{g}) q(\bm{e}) q(\bm{\Lambda}) q(\bm{\Gamma}) q(\bm{\Delta}) q(\sigma^2).
\end{eqnarray*}

For our model, the resulting approximate posterior distribution of each factor follows the same distribution as the corresponding prior. Table \ref{table:priorVar} presents the assumed prior distributions as well as the approximations of the posterior distribution, where $\mathcal{N}$, $\mathcal{N^+}$ and $\mathcal{G}$ denote the normal, truncated normal and gamma distributions, respectively. For brevity, we do not list here the variational updates of all these parameters. We provide the full mathematical derivation in Appendix \ref{VU}.

A key point in variational inference is the dependence of the estimates on the initial values of the optimization \citep{rossi2019good, airoldi2008mixed, lim2007variational}. In our variational approach, the initial values proved to be especially important, perhaps due to the dependence on the parameters of the bilinear term, which is ignored by the coordinate ascent. In view of this, our way to get around this problem was to initialize the $\Theta$ set in the algorithm with the frequentist estimates of the parameters, then we recursively update the variational parameters $\mu_{q(\mu)}$, $\Sigma_{q(\mu)}$, $\mu_{q(g_i)}$, $\Sigma_{q(g_i)}$, $\mu_{q(e_j)}$, $\Sigma_{q(e_j)}$, $\mu_{q(\lambda_q)}$, $\Sigma_{q(\lambda_q)}$, $\mu_{q(\gamma_i)}$, $\Sigma_{q(\gamma_i)}$, $\mu_{q(\delta_j)}$, $\Sigma_{q(\delta_j)}$, $a_{q(\sigma^{-2})}$, and $b_{q(\sigma^{-2})}$, for all $i, j, q$, until convergence. For more discussion of this initialization see Section \ref{results}.

\begin{table}[h!]
\centering
\rowcolors{2}{gray!25}{white}
\def\arraystretch{2}\tabcolsep=10pt
\caption{Overview of prior distributions and the respective variational distributions of each parameter of the model.}
\begin{tabular}{ccc}
\hline
Parameter      & Prior Distribution                                                     & Variational Distribution                                          \\ \hline
$\mu$          & $\mathcal{N}(\mu_\mu ; \sigma^2_\mu)$                                  & $\mathcal{N}(\mu_{q(\mu)}, \Sigma_{q(\mu)})$                      \\
$\bm{g}$       & $\prodi \mathcal{N}(0 ; \sigma^2_g)$                                   & $\prodi \mathcal{N}(\mu_{q(g_i)}, \Sigma_{q(g_i)})$               \\
$\bm{e}$       & $\prodj \mathcal{N}(0 ; \sigma^2_e)$                                   & $\prodj \mathcal{N}(\mu_{q(e_j)}, \Sigma_{q(e_j)})$               \\
$\bm{\lambda}$ & $\text{ordered sample of } \prodq \mathcal{N}^+(0 ; \sigma^2_\lambda)$ & $\prodq \mathcal{N}^+(\mu_{q(\lambda_q)}, \Sigma_{q(\lambda_q)})$ \\
$\bm{\gamma}$  & $\mathcal{N}^+(0 ; 1) \prodq \prod\limits_{i=2}^{I} \mathcal{N}(0 ; 1)$     & $\prodi \mathcal{N}(\mu_{q(\gamma_i)}, \Sigma_{q(\gamma_i)})$     \\
$\bm{\delta}$  & $\prodq\prodj \mathcal{N}(0 ; 1)$                                      & $\prodj \mathcal{N}(\mu_{q(\delta_j)}, \Sigma_{q(\delta_j)})$     \\
$\sigma^{-2}$  & $\mathcal{G}(a ; b)$                                                   & $\mathcal{G}(a_{q(\sigma^{-2})}, b_{q(\sigma^{-2})})$             \\ \hline
\end{tabular}
\label{table:priorVar}
\end{table}

\section{Results} \label{results}

\subsection{Simulation Study}

In this section we describe the simulation study to investigate the performance of the proposed algorithm. The data were generated as follows: 
\begin{enumerate}
    \item The number of environments and genotypes were $I \in \{6, 10, 12, 25, 50, 100, 200\}$ and $J \in \{10,12,20,30, 50, 100\}$, respectively, and we used $Q \in \{1,2\}$.
    \item We simulated the additive term as $\mu\sim\mathcal{N}(90, 10)$, $g_i\sim\mathcal{N}(0, \sigma_g^2)$ and $e_j\sim\mathcal{N}(0,\sigma_e^2)$. We fixed the standard deviations $\sigma_g^2 = 10$ and $\sigma_e^2 = 10$.
    \item For the multiplicative term, we fixed $\lambda = \{12, 20, 25\}$, and after applying orthonormalization, we obtained the matrices $\bm{\Gamma}$ and $\bm{\Delta}$.
    \item We simulated the values $y_{ij}$ from $\mathcal{N}(\mu +g_{i}+e_{j}+\sum_{q=1}^{Q} \lambda_{q} \gamma_{iq} \delta_{jq}, \sigma_y^2)$, where we fixed $\sigma_y^2=1$.
\end{enumerate}
  
\begin{figure}[H]
\centering
        \includegraphics[width=\columnwidth]{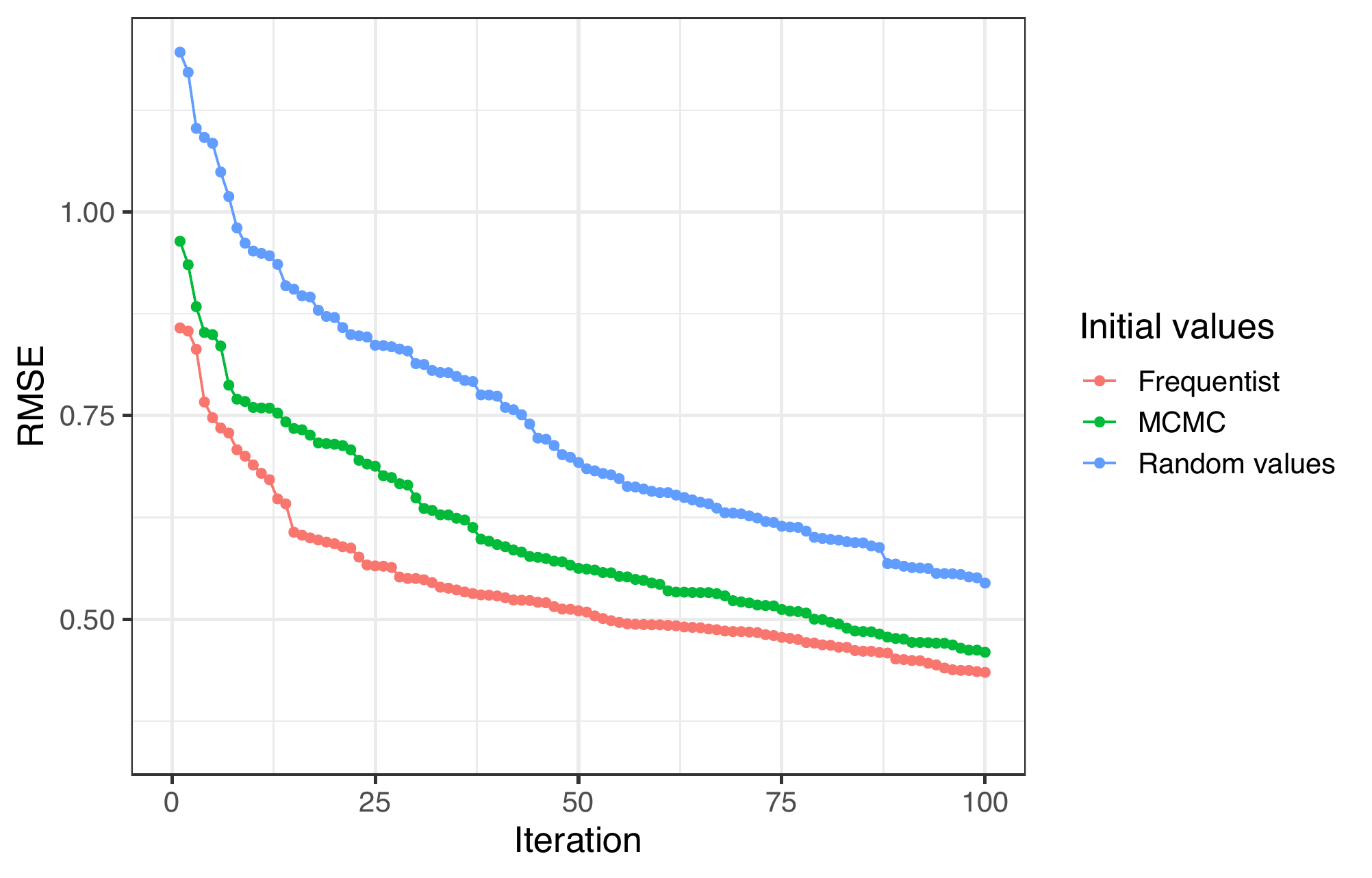}
			\caption{Comparison of the RMSE of the VI in each iteration considering three different initialization: random initialization, frequentist estimates and MCMC estimates.}
        \label{fig:initialrmse}
\end{figure}

We performed several experiments to determine how the Root Mean Square Error (RMSE) is influenced by the initialization of the $\bm{\Theta}$ vector in the algorithm.  To achieve this, we altered the initial values of the $\bm{\Theta}$ vector in three different scenarios. That is, for each scenario we supplied as initial values: (a) random values from a normal distribution; (b) frequentist estimates from the AMMI model; and (c) Bayesian estimates of the model, considering 25\% of the simulated data. The simulation scenario used in the experiment was $\lambda = 12$, $I = 25$, $J = 12$ and $Q=1$. Looking at Figure \ref{fig:initialrmse}, we see that frequentist and Bayesian estimates have better performance (lower RMSE), but as MCMC has a higher computational cost, we consider the former to be the best option to initiliase the VI algorithm.
  
  We evaluated the performance by calculating the accuracy of the predicted values, computational time and compatibility of the estimates compared to those obtained via MCMC. We used RMSE as a measure to assess the quality of the estimates in the predicted values. In the evaluation of the criteria for accuracy and comparison of estimates, we used the results from the implementation of MCMC using the package `R2jags' \citep{plummer2003jags, su2012r2jags}. The computational time was measured in minutes, and, for comparison purposes, we implemented MCMC via a Gibbs sampling scheme. We ran 4 chains, with 6,000 iterations each, with a burn-in period of 1,000 iterations, thus 20000 simulations. Convergence was observed by monitoring the trace-plot and Gelman and Rubin's convergence diagnostic at 95\% \citep{gelman1992inference}.  All  algorithms were implemented in R \citep{R}, on a MacBook Pro 1.4GHz Quad-Core Intel Core i5 with 8GB memory.

\begin{figure}[H]
\begin{subfigure}[t]{.5\textwidth}\centering
  \includegraphics[width=\columnwidth]{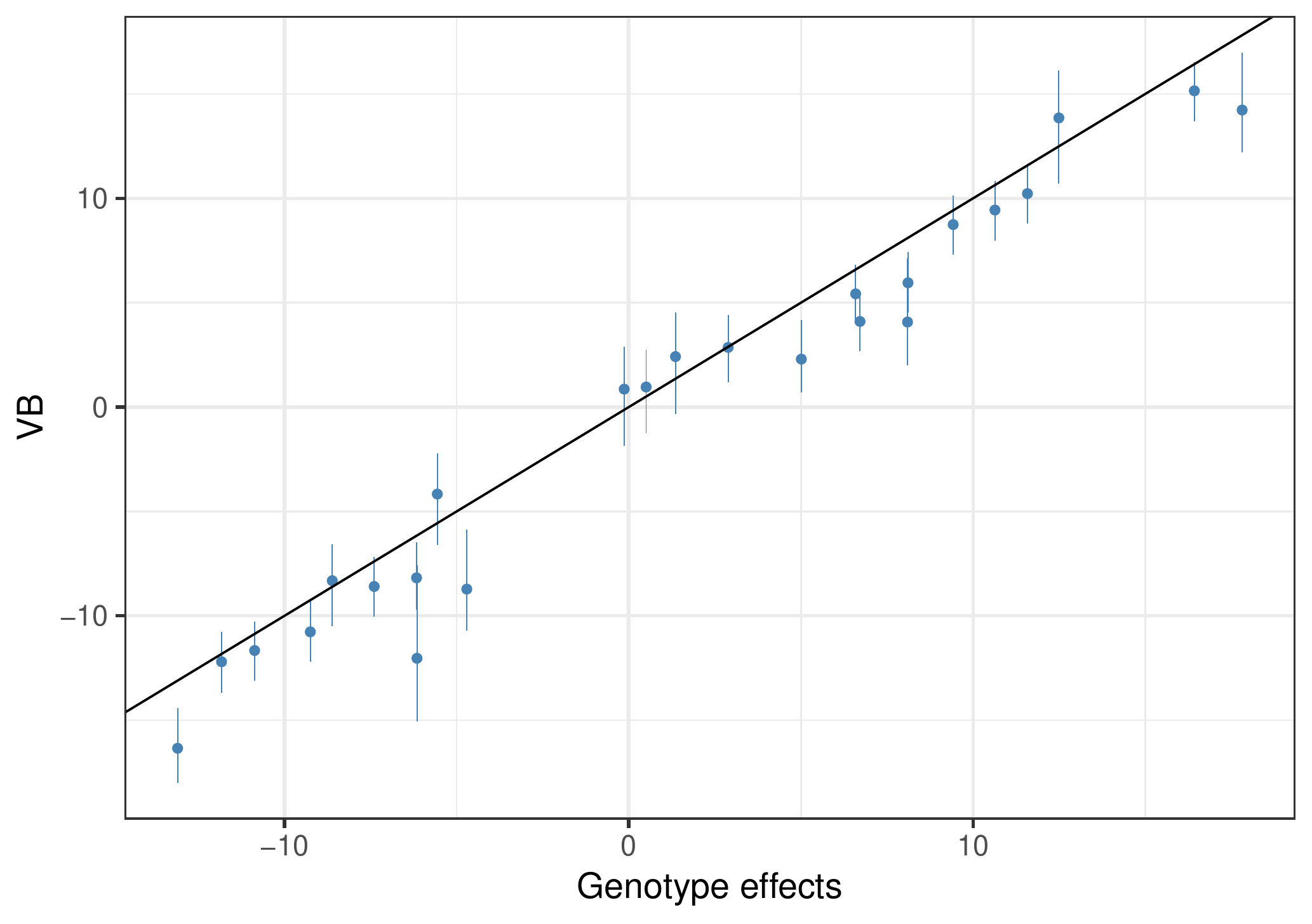}
  \caption{True (horizontal axis) versus estimated (vertical axis) values of genotype effects.}
  \label{fig: gen}
\end{subfigure}%
\begin{subfigure}[t]{.5\textwidth}\centering
  \includegraphics[width=\columnwidth]{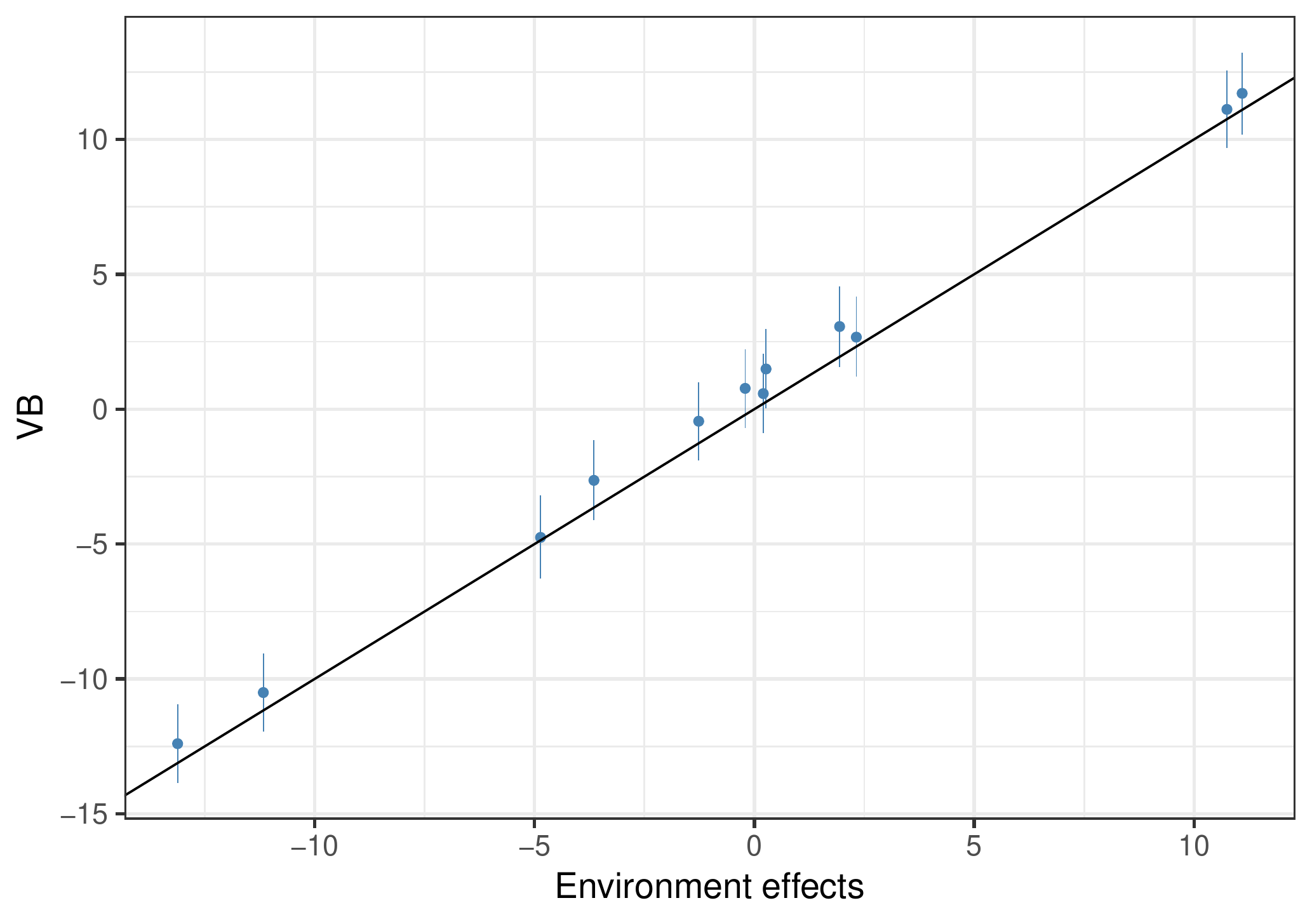}
  \caption{True (horizontal axis) versus estimated (vertical axis) of environment effects.}
\end{subfigure}
\caption{Variational estimates of genotype and environment effects considering $Q = 1$, $\lambda = 20$, $I = 25$, $J = 12$, and the the respective quantile intervals, 5\% and 95\%.}
    \label{fig:maineffsim}
\end{figure}

The main effects parameters and the mean behave well, in the sense that the respective estimates converge to their true values, regardless of the Q value taken. The variational distributions of these parameters are in line with those obtained via MCMC. The variational estimates of the main effects of genotypes and environments versus the true values are shown in Figure \ref{fig:maineffsim} as well as the respective subsequent MCMC, considering Q = 1, $\lambda = 20$, 25 genotypes and 12 environments. 

\begin{figure}[H]
\begin{subfigure}[t]{.5\textwidth}\centering
  \includegraphics[width=\columnwidth]{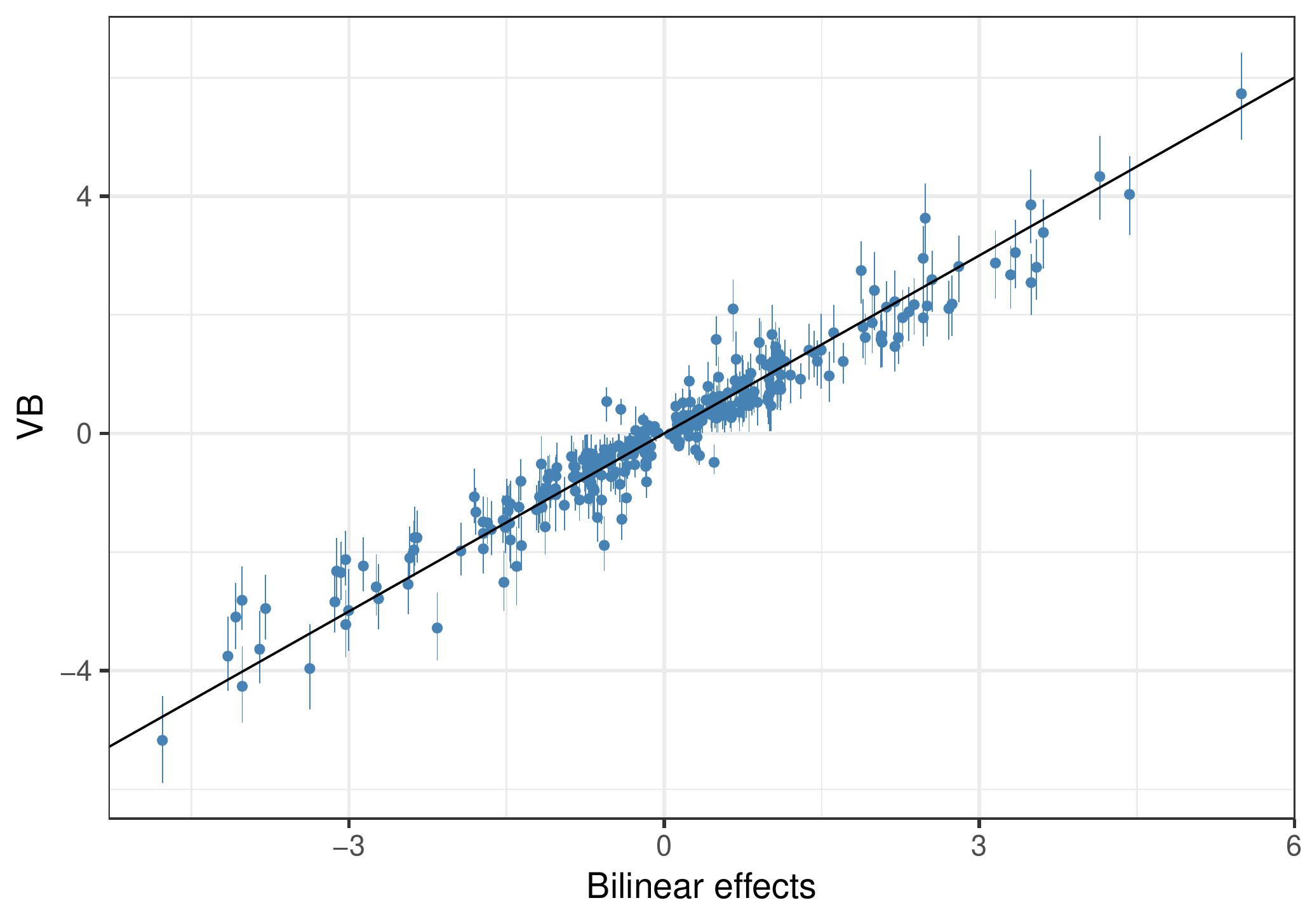}
  \caption{True bilinear effects (horizontal axis) versus variational estimates (vertical axis).}
  \label{fig: blin}
\end{subfigure}%
\begin{subfigure}[t]{.5\textwidth}\centering
  \includegraphics[width=\columnwidth]{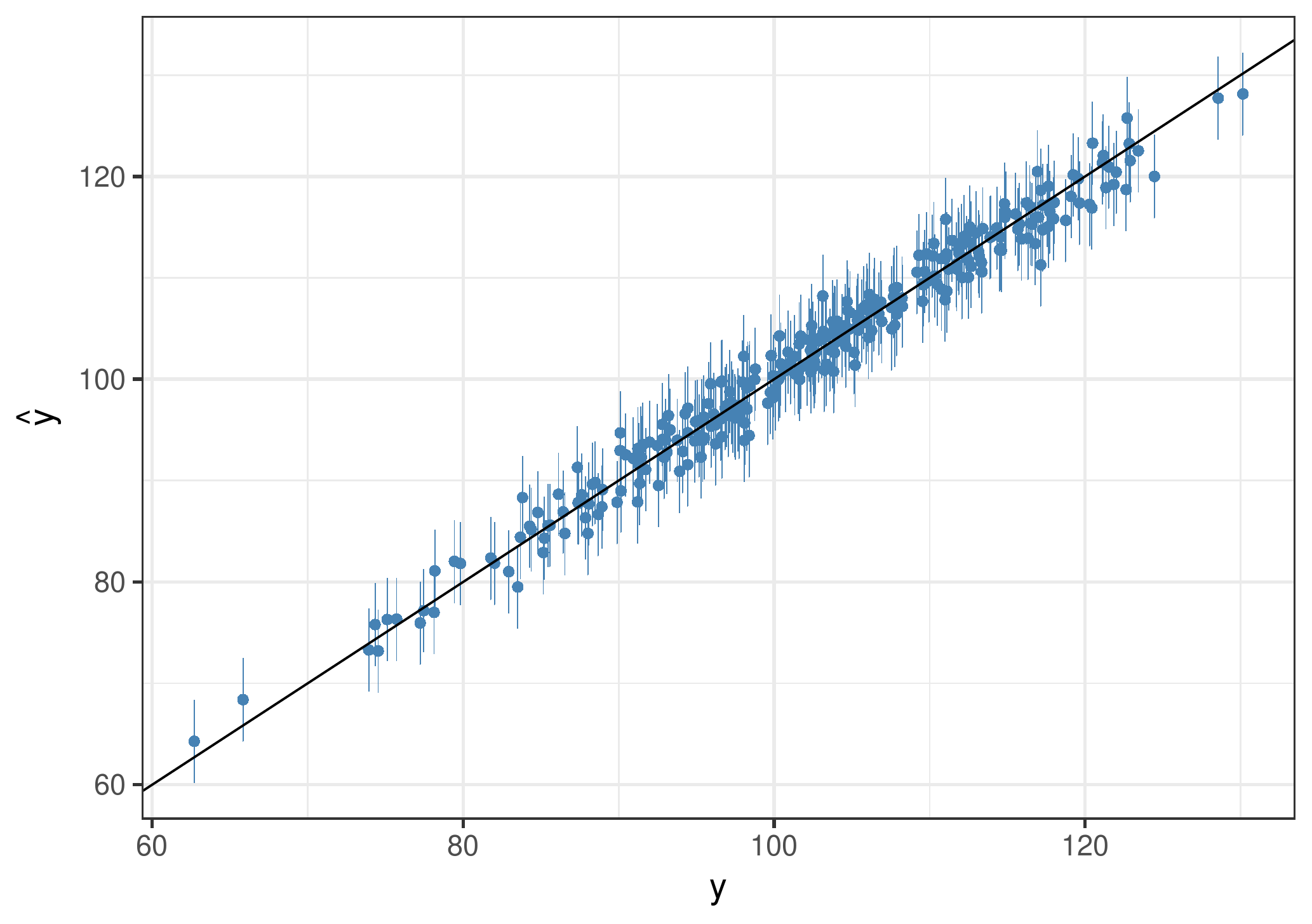}
  \caption{True values of $y$ (horizontal axis) versus predicted values (vertical axis).}
\end{subfigure}
\caption{Variational estimates of bilinear term and predicted values considering $Q = 1$, $\lambda = 20$, $I = 25$, $J = 12$, and the the respective quantile ranges, 5\% and 95\%.}
    \label{fig:blin figures}
\end{figure}

Although the results for the main effects behave well, our main interest is to analyze the quality of the estimation of the bilinear interaction term. To achieve this, we analyzed three scenarios in particular, one when $\lambda$ was close to zero, another when $\lambda$ was close to 20, and finally when $\lambda$ was close to 40, with $Q=1$ for each scenario. The first case performs the worst, as the model cannot capture the interaction and, consequently, the estimates are poor. However, as $\lambda$ grows,  the simulation results showed that the VI algorithm is able to estimate the bilinear parameters reasonably well, see Figure \ref{fig: blin}.

\begin{figure}[H]
    \begin{subfigure}[b]{0.45\textwidth}
        \includegraphics[width=\columnwidth]{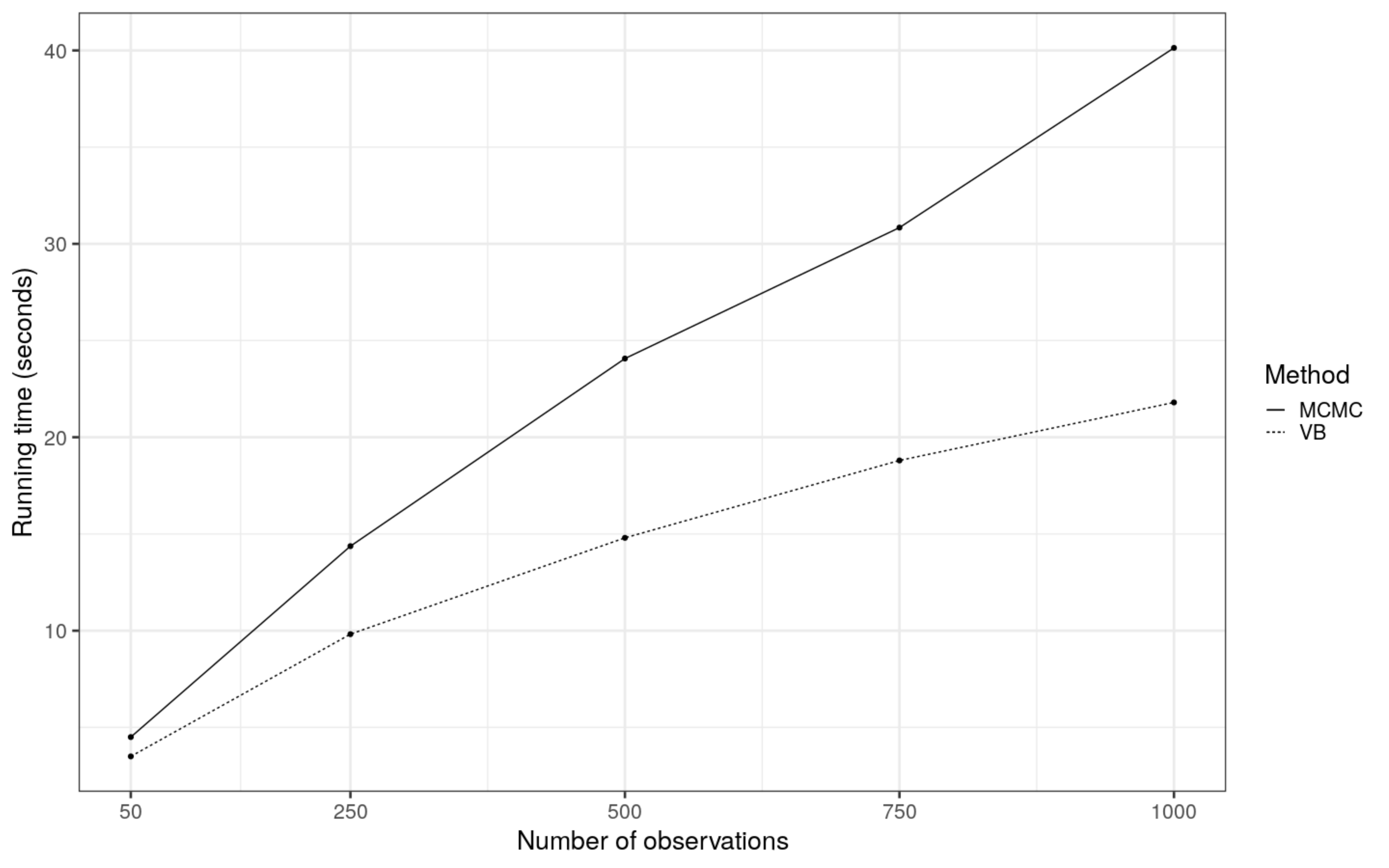}
			\caption{Comparison of simulation times for y = $\{100, 250, 500, 1000\}$.}
        \label{fig:t1}
    \end{subfigure}
    \hfill
    \begin{subfigure}[b]{0.45\textwidth}
        \includegraphics[width=\columnwidth]{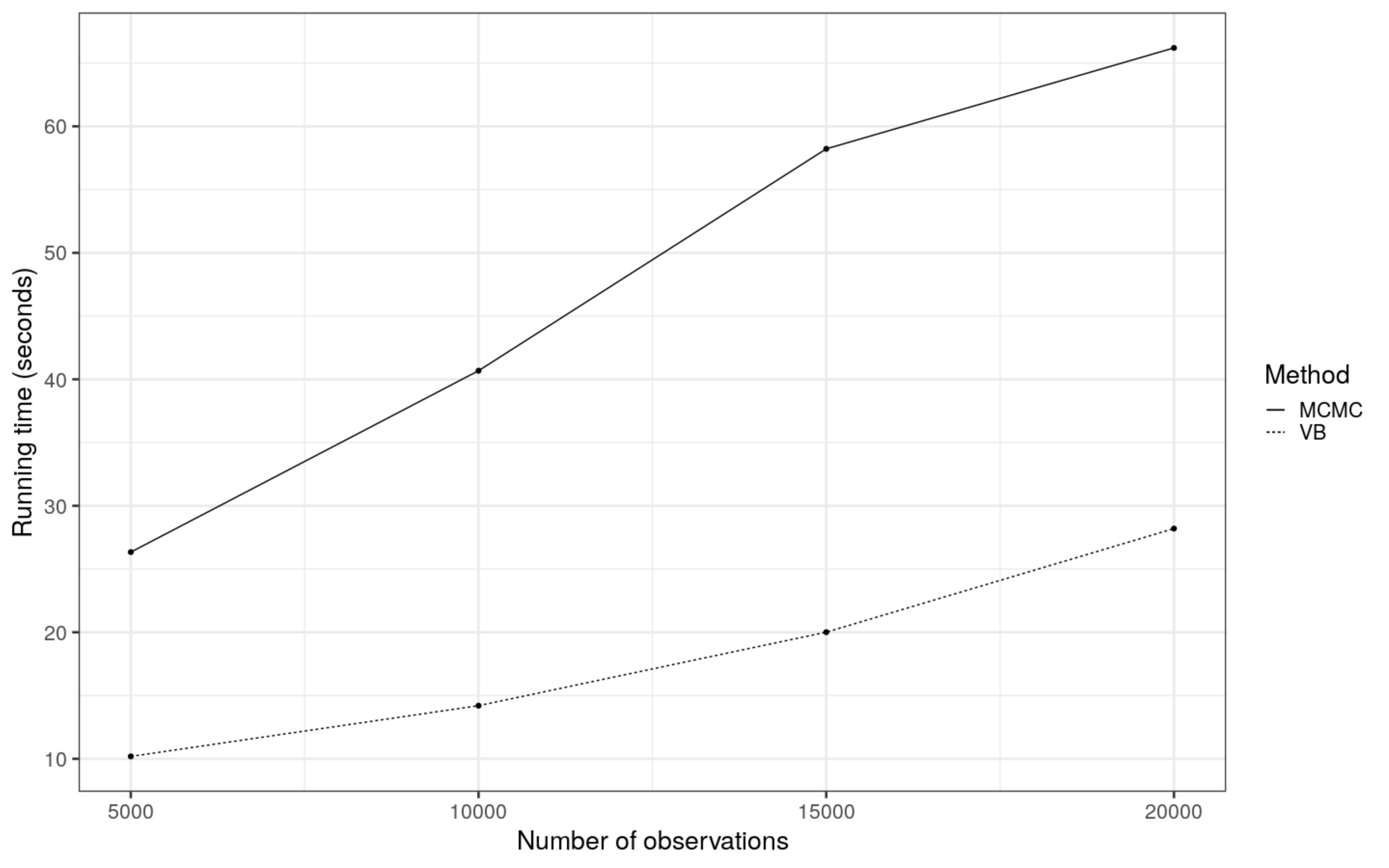}
        \caption{Comparison of simulation times for y = $\{5000, 10000, 15000, 20000\}$.}
        \label{fig:t2}
    \end{subfigure}
    \caption{Computational fitting time of AMMI model VI versus MCMC. For smaller experiments where the number of genotypes/environments is in the 10s then the time difference is small. For large experiments where $I,J>100$ then the benefits of the VI model are clearer. }
    \label{fig:time figures}
\end{figure}

To evaluate the computational performance of the VI algorithm, we considered different scenarios, and separated them into two different groups: smaller versus larger sample sizes, which consisted of $\{100, 250, 500, 1000\}$  and $\{5000, 10000, 15000, 20000\}$ observations respectively. When the number of observations is smaller ($I\times J \leq 1000$) the computational cost of MCMC and VI are similar (Figure \ref{fig:t1}). However, when there is an increase in the number of genotypes and environments ($I\times J > 5000$), it is possible to observe a more significant difference between the two procedures (Figure \ref{fig:t2}). The value of $Q$ is also naturally changes the computational  time of the algorithms, and as it grows, there is a change in the unit of time measurement (that is, from seconds to hours as Q increases).

\subsection{Real Data Set}

We now illustrate our methods applied to a real data set from the Horizon2020 EU InnoVar project (\url{www.h2020innovar.eu}) that aims to build new cultivation tools from genomic, phenomic and environmental data. For this study, we consider data spanning ten years (2010 - 2019) concerning the production of a common species of wheat called \textit{Triticum aestivum L.}, in Ireland, with the response being the yield of wheat measured in tonnes per hectare. The data were supplied by the Irish Department of Agriculture, Food, and Marine. The experiments were conducted using a block design with four replicates, with the yield averaged across the replicates.  For our study, we considered a single data set with all years, taking the averages of the years of genotypes and environments, resulting in a final data set containing 85 genotypes, 17 environments, and a total of 810 observations.

Initially, our objective is to perform a comparison of VI to MCMC, evaluating the computational cost of the two methods, contrasting the posterior distributions and the accuracy of the predictions. Then, we use the posterior variational distributions to make inferences about the genotypes and environments under study, identifying which genotypes perform better in each environment. In both algorithms, we fit the model by taking $Q = 1$ and $Q = 2$. 

\begin{figure}[h]
\centering
        \includegraphics[width=\columnwidth]{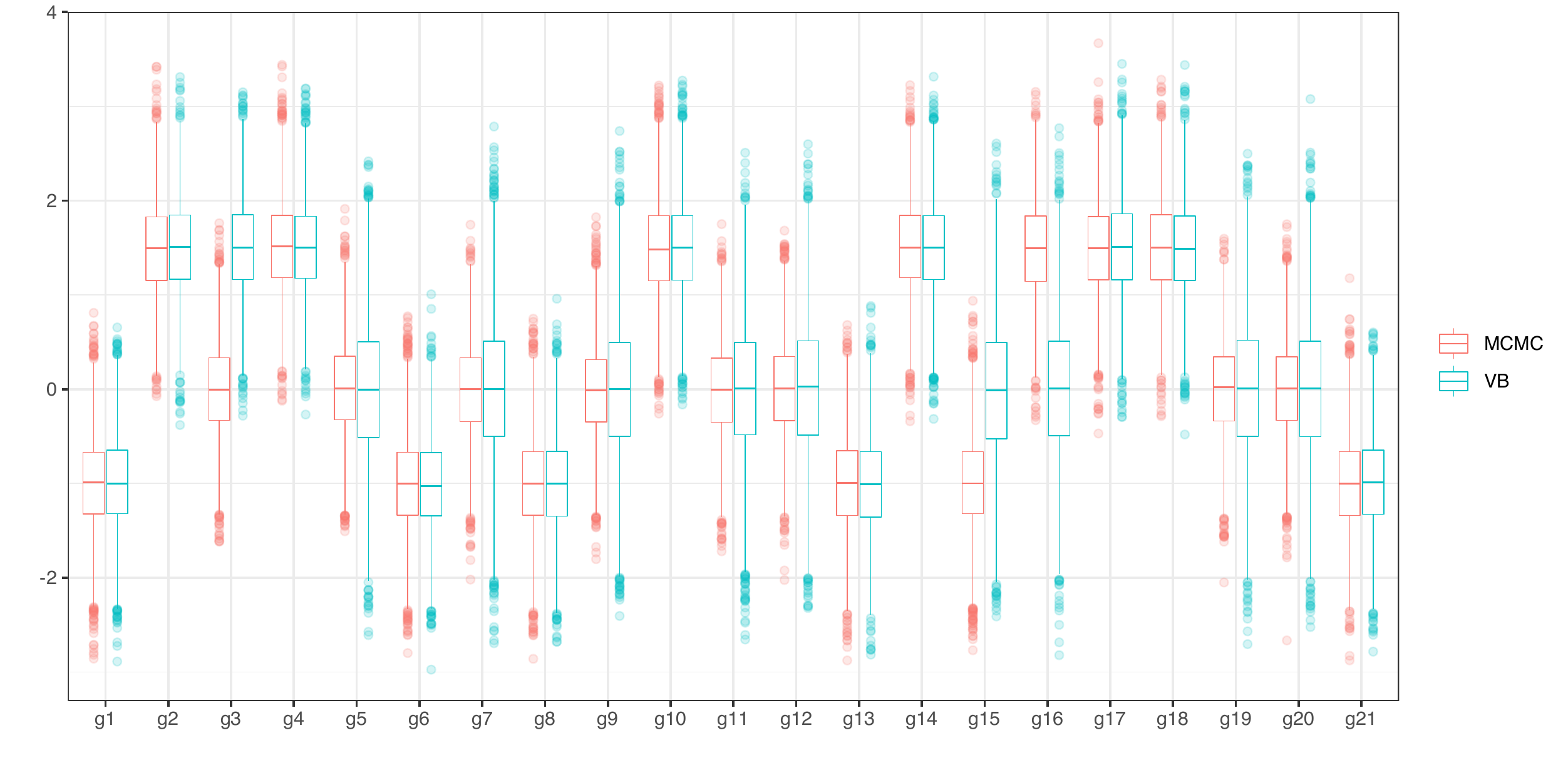}
			\caption{Posterior distributions of the first 21 genotypes obtained via VI and MCMC (Q=1).}
        \label{fig:compmcmcvb}
\end{figure}

In Figure \ref{fig:compmcmcvb} we present the comparison between the posterior distributions of the genotype term obtained by the two approaches, for the case where Q = 1. We can see that the distributions have a very similar behavior, being around the same mean. On the other hand, the VI method has a higher variance in some cases. The results for the other parameters and for Q = 2 are presented in the Appendix \ref{MSR}. 

In terms of computational time, the two methods were equivalent for Q = 1, whereas for Q = 2, VI is faster. Regarding the accuracy of the predictions, calculated in sample, although it is known that the VI approach is less accurate than MCMC, in the data set under study the VI was quite satisfactory, since it had an RMSE of 0.58 for $Q = 1$, and 0.54 for $Q = 2$, while for the MCMC it was 0.53 and 0.52, respectively.

The second point to be addressed concerns the effect of genotypes in respective environments. \cite{josse2014another} discuss how the Bayesian methodology could be used to provide additional insights into the analysis of \verb+GxE+ interactions. For example, which genotype has the best performance across environments, and which genotype has the best performance in a given environment. In the classical methodology, this type of question is answered using a biplot \citep{gabriel1978least}. However, some authors have already discussed how careful the researcher should be when using this tool. \cite{josse2014another} presents a graphical way in which credibility boxes are created using the quantiles of the posterior distributions. \cite{sarti2021bayesian} propose a heatmap plot to observe which genotype and environment are best for producing wheat. This plot, when compared with the traditional biplot, provides more complete and objective information about which genotypes and environments interact most effectively, in an intuitive and interpretable way.

\begin{figure}[H]
    \begin{subfigure}[b]{0.45\columnwidth}
        \includegraphics[width=\textwidth]{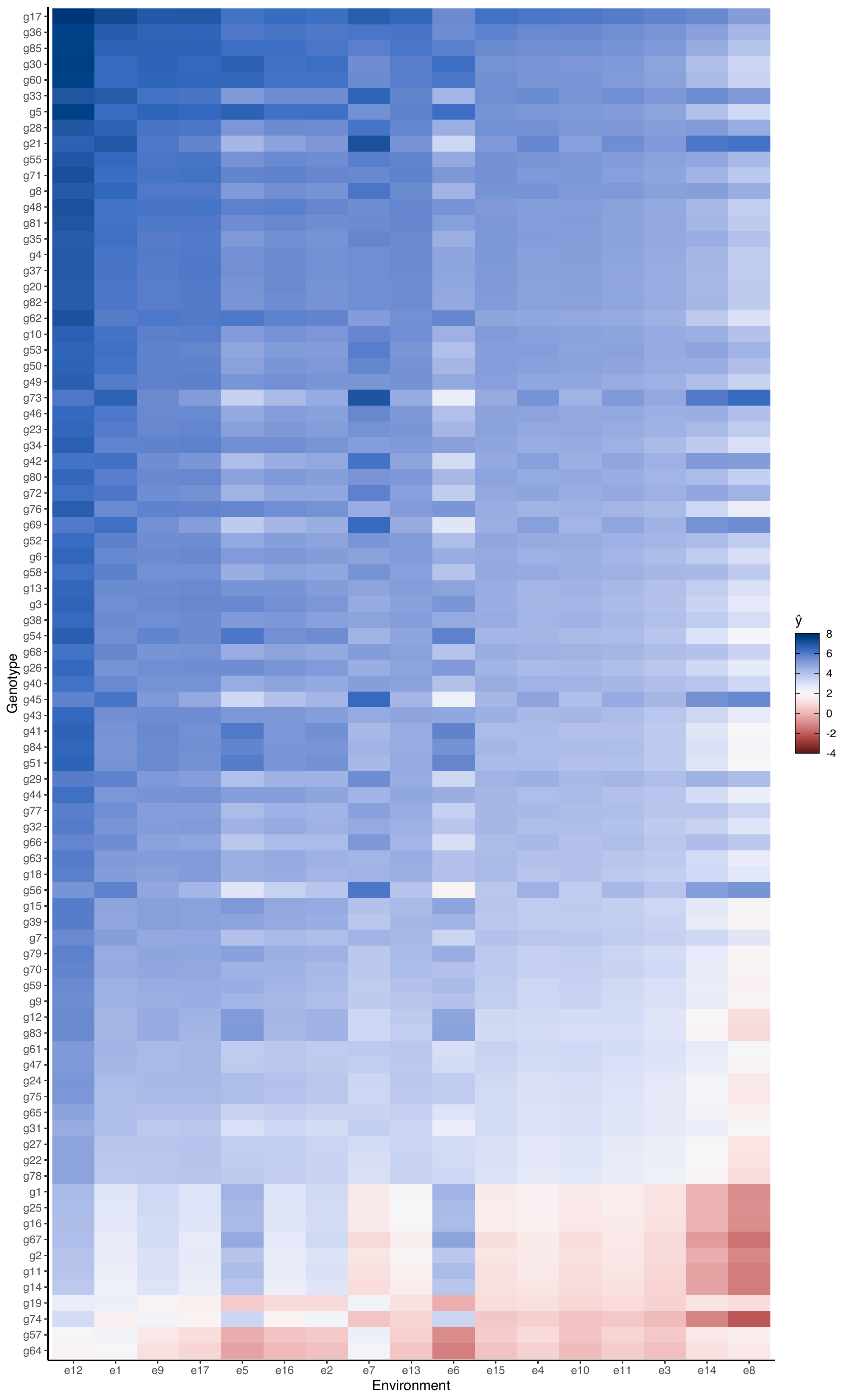}
			\caption{5\% quantile.}
        \label{fig:h1}
    \end{subfigure}
    \hfill
    \begin{subfigure}[b]{0.45\columnwidth}
        \includegraphics[width=\textwidth]{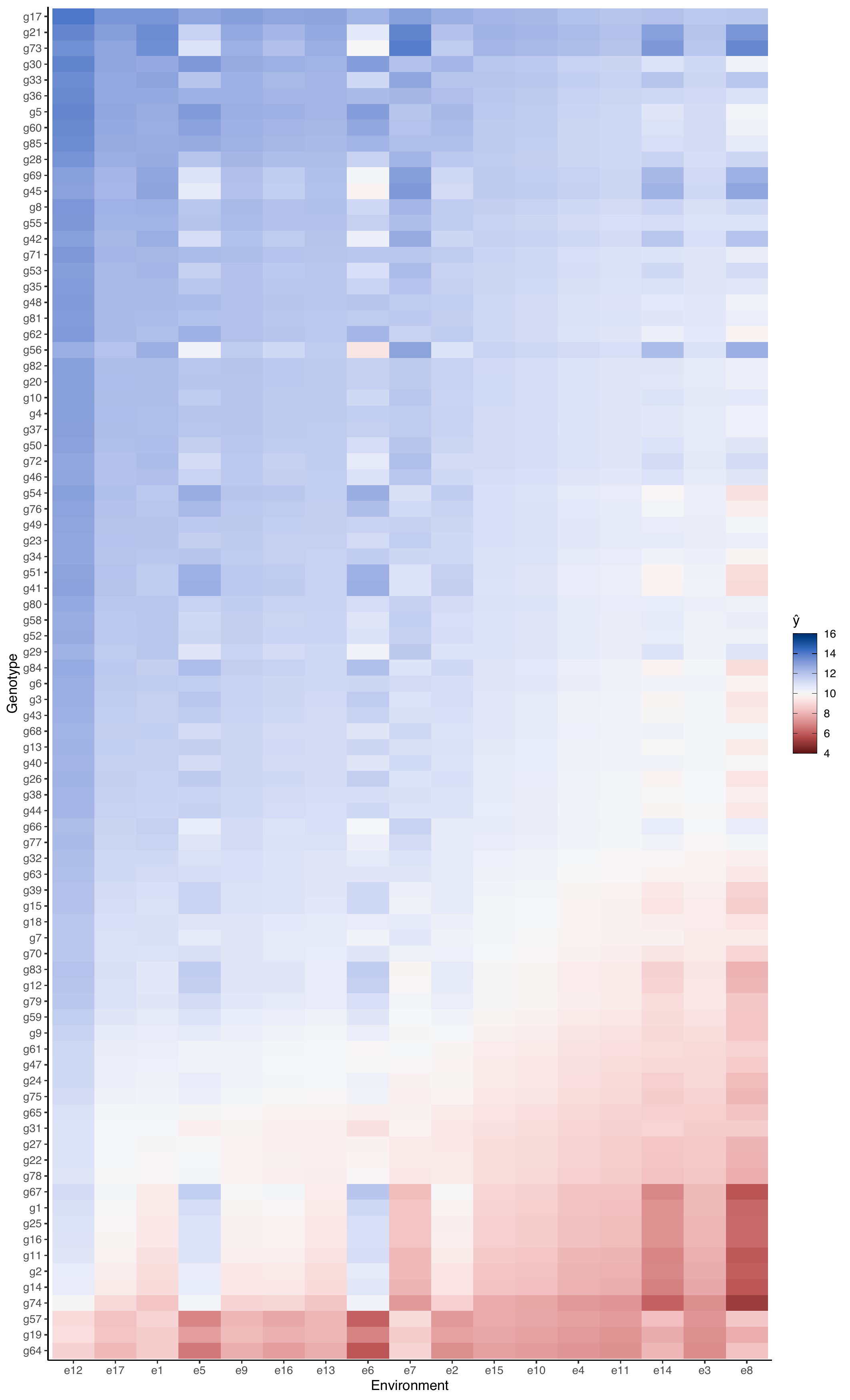}
        \caption{50\% quantile.}
        \label{fig:h2}
    \end{subfigure}
    \hfill
    \begin{subfigure}[b]{0.45\columnwidth}
        \includegraphics[width=\textwidth]{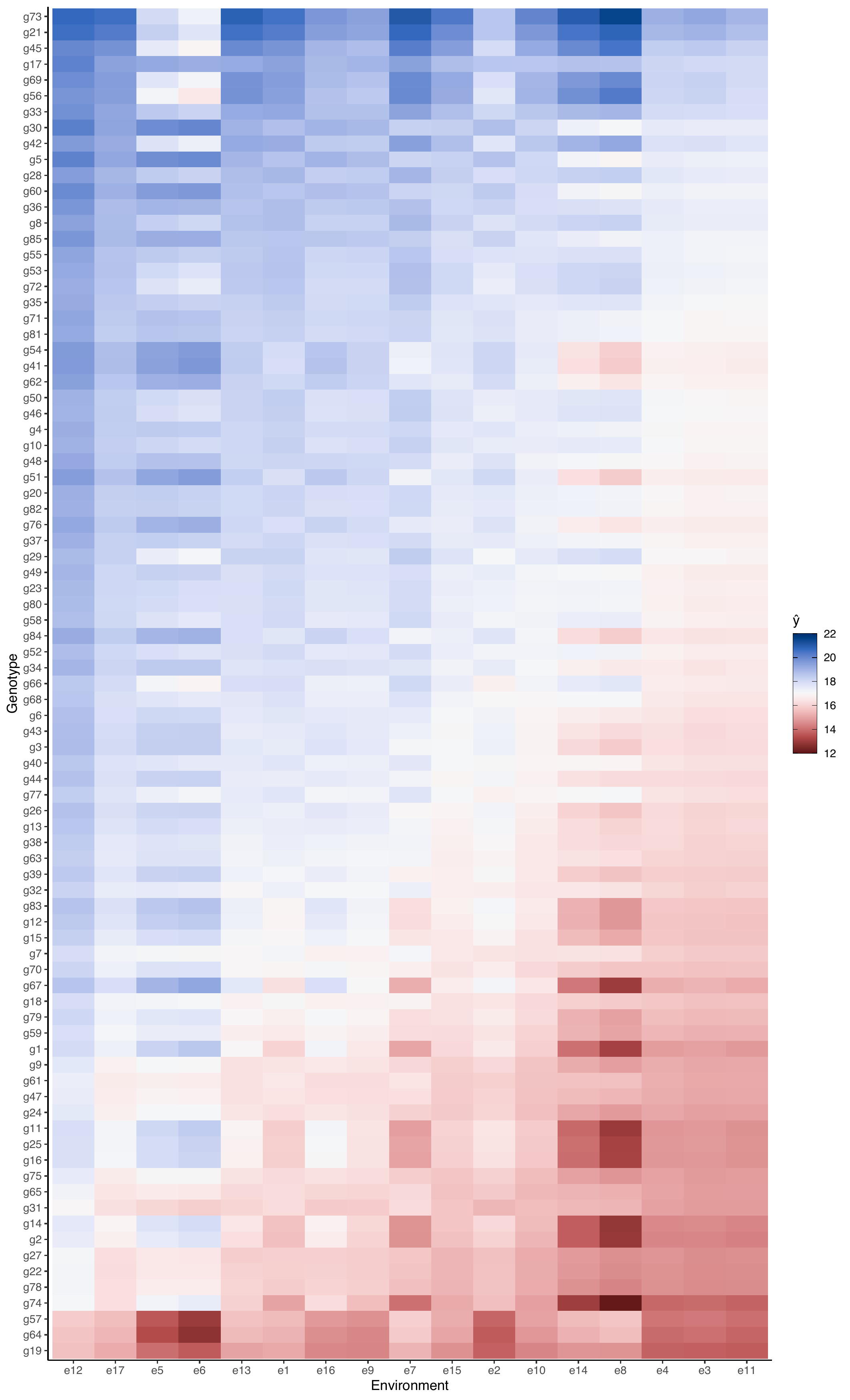}
        \caption{95\% quantile.}
        \label{fig:h3}
    \end{subfigure}
    \caption{Quantiles 5\%, 50\% and 95\% of the predicted wheat yields, from the posterior variational distribution ($Q = 1$).}
    \label{fig:heatmaps}
\end{figure}

Figure \ref{fig:heatmaps} shows the visualization proposed by \cite{sarti2021bayesian}, considering the variational posterior distributions setting $Q = 1$. Figure \ref{fig:h2} shows that environment $e_{12}$ interacts better with genotypes $g_{17}$ and $g_{21}$. Additionally, genotypes $g_{21}$ and $g_{73}$  have better production in environment $e_7$. In contrast, genotype $g_{74}$ has the lowest production in environment $e_8$. We can also see that environment $e_8$ has the lowest wheat yields, while environments $e_{12}$, $e_{17}$, $e_{1}$, $e_{5}$, and $e_{9}$ have the best yields. Genotypes $g_{17}$, $g_{21}$, and $g_{73}$ performed well in practically all environments in which it was present, while genotypes $g_{57}$, $g_{19}$ and $g_{64}$ showed lower values in all environments. Figures \ref{fig:h1} and \ref{fig:h3} show the 5\% and 95\% quantiles, which are uncertainties associated with the predicted yields. Only in environments $e_7$, $e_2$, $e_4$, and $e_{11}$ were all genotypes present. The results observed setting $Q = 2$ are similar and presented in the Appendix \ref{MSR}.

\section{Discussion}\label{discussion}

In this work we performed variational inference on the Additive Main Effects and Multiplicative Interaction Effect models, which is widely used in analyzing \verb+GxE+ interactions. Our main contribution was formulating an efficient variational
approximation scheme for inference following the priors suggested by \cite{josse2014another} in order to meet the model constraints and obtain a computationally faster algorithm.

As shown in Section \ref{results}, in contrast to other Bayesian methods already presented in the literature, as well as the results presented by \cite{josse2014another}, we found the inferential approach to perform well in several simulation scenarios, considering both small and larger sample sizes. The computational time proved to be superior when compared to both the algorithm when run in JAGS and Gibbs sampling. This performance was carried over into the real InnoVar data set where predictive performance was similar to MCMC whilst being roughly two times faster.

Given the results, we believe that the variational version of AMMI model is competitive due to our method's simplicity and vast speed improvements. Further improvements to the model might be made in terms of finding better initialisations of the approach. 

We hope that others find to be useful when an AMMI model is used to fit \verb+GxE+ interactions, particularly for large data sets.

\begin{acks}[Acknowledgments]
Antônia A. L. dos Santos, Andrew Parnell, and Danilo Sarti received funding for their work from the European Union’s Horizon 2020 research and innovation programme under grant agreement No 818144. In addition Andrew Parnell’s work was supported by: a Science Foundation Ireland Career Development Award (17/CDA/4695); an investigator award (16/IA/4520); a Marine Research Programme funded by the Irish Government, co-financed by the European Regional Development Fund (Grant-Aid Agreement No. PBA/CC/18/01); SFI Centre for Research Training in Foundations of Data Science 18/CRT/6049, and SFI Research Centre awards I-Form 16/RC/3872 and Insight 12/RC/2289\_P2. For the purpose of Open Access, the author has applied a CC BY public copyright licence to any Author Accepted Manuscript version arising from this submission.
\end{acks}

\bibliographystyle{imsart-nameyear} 
\bibliography{references}

\begin{appendix}

\section{Additional Results}\label{MSR}

\begin{figure}[H]
    \begin{subfigure}[b]{0.39\columnwidth}
        \includegraphics[width=\textwidth]{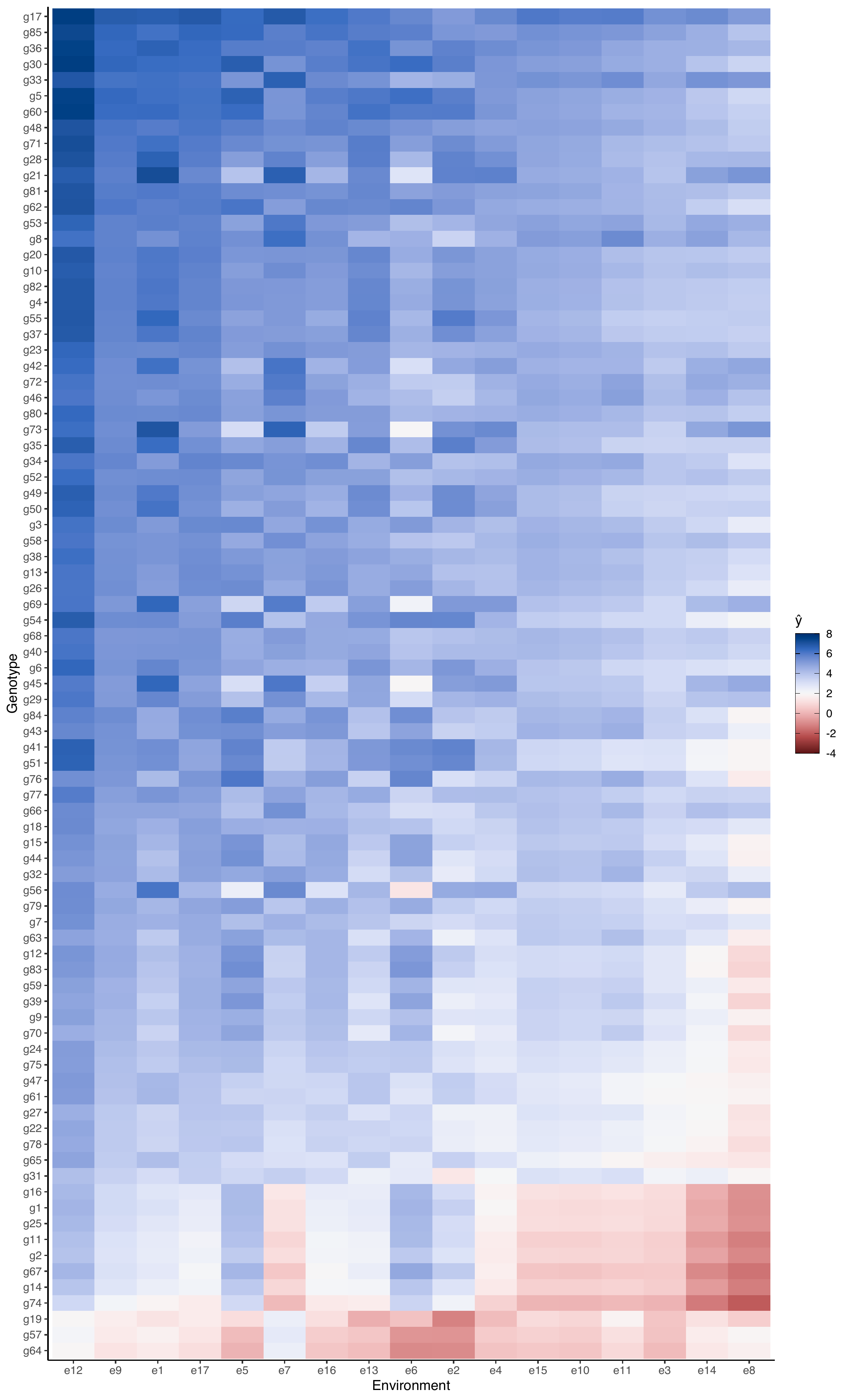}
			\caption{5\% quantile.}
        \label{fig:h1-2}
    \end{subfigure}
    \hfill
    \begin{subfigure}[b]{0.39\columnwidth}
        \includegraphics[width=\textwidth]{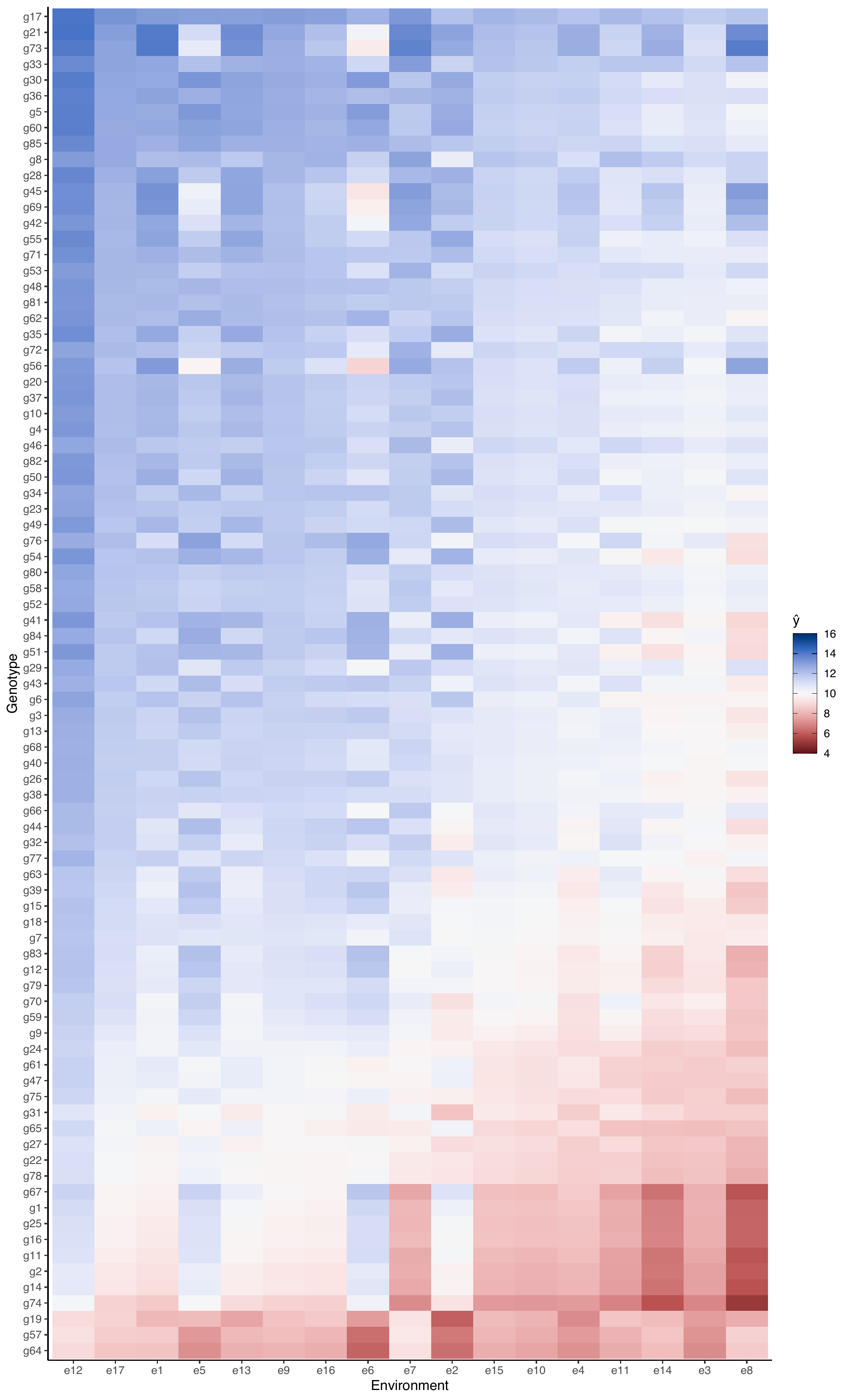}
        \caption{50\% quantile.}
        \label{fig:h2-2}
    \end{subfigure}
    \hfill
    \begin{subfigure}[b]{0.39\columnwidth}
        \includegraphics[width=\textwidth]{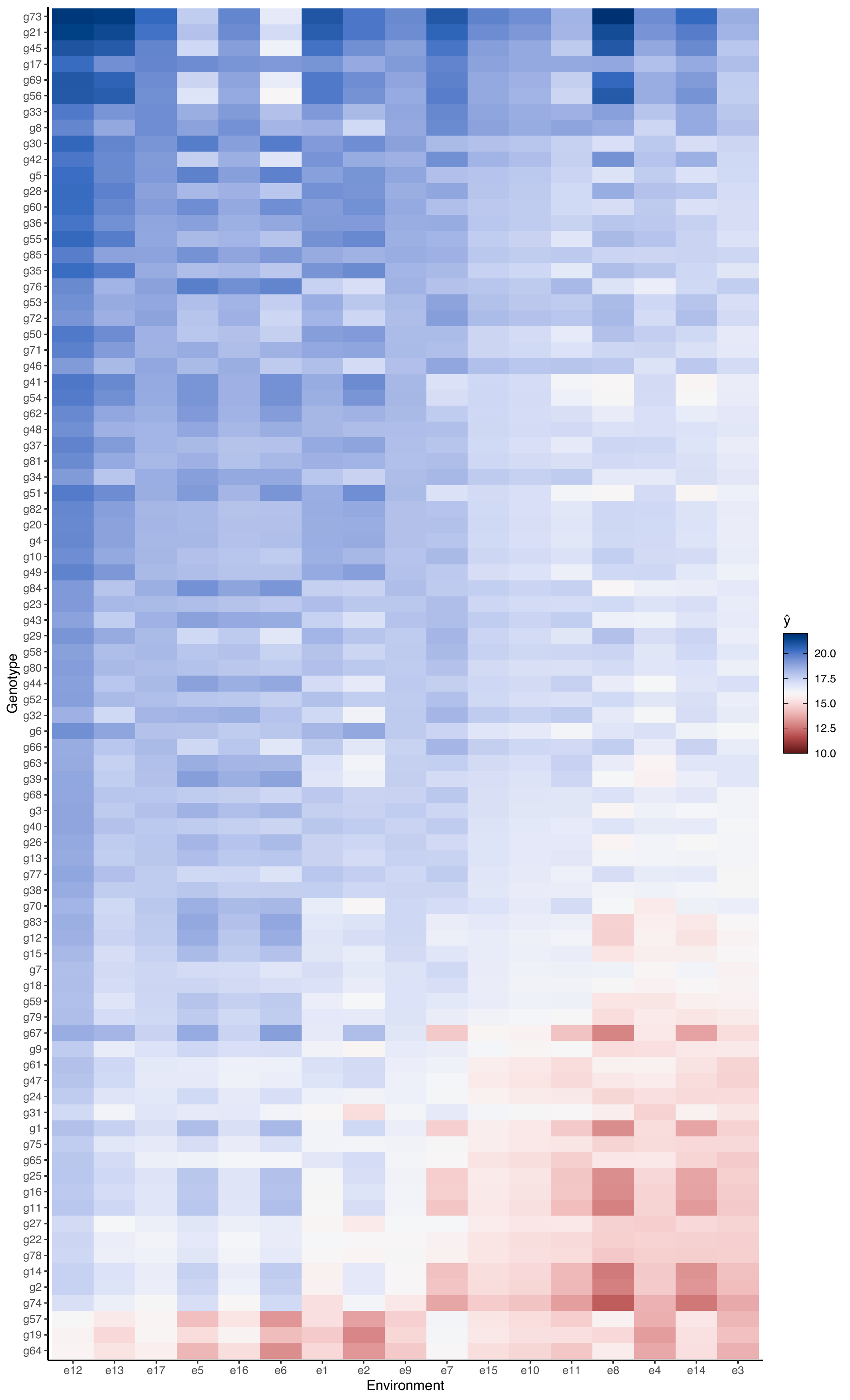}
        \caption{95\% quantile.}
        \label{fig:h3-2}
    \end{subfigure}
    \caption{Quantiles 5\%, 50\% and 95\% of the predicted wheat yields, from the posterior variational distribution ($Q = 2$). }
    \label{fig:heatmaps2}
\end{figure}

\begin{figure}[h]
\centering
        \includegraphics[width=\columnwidth]{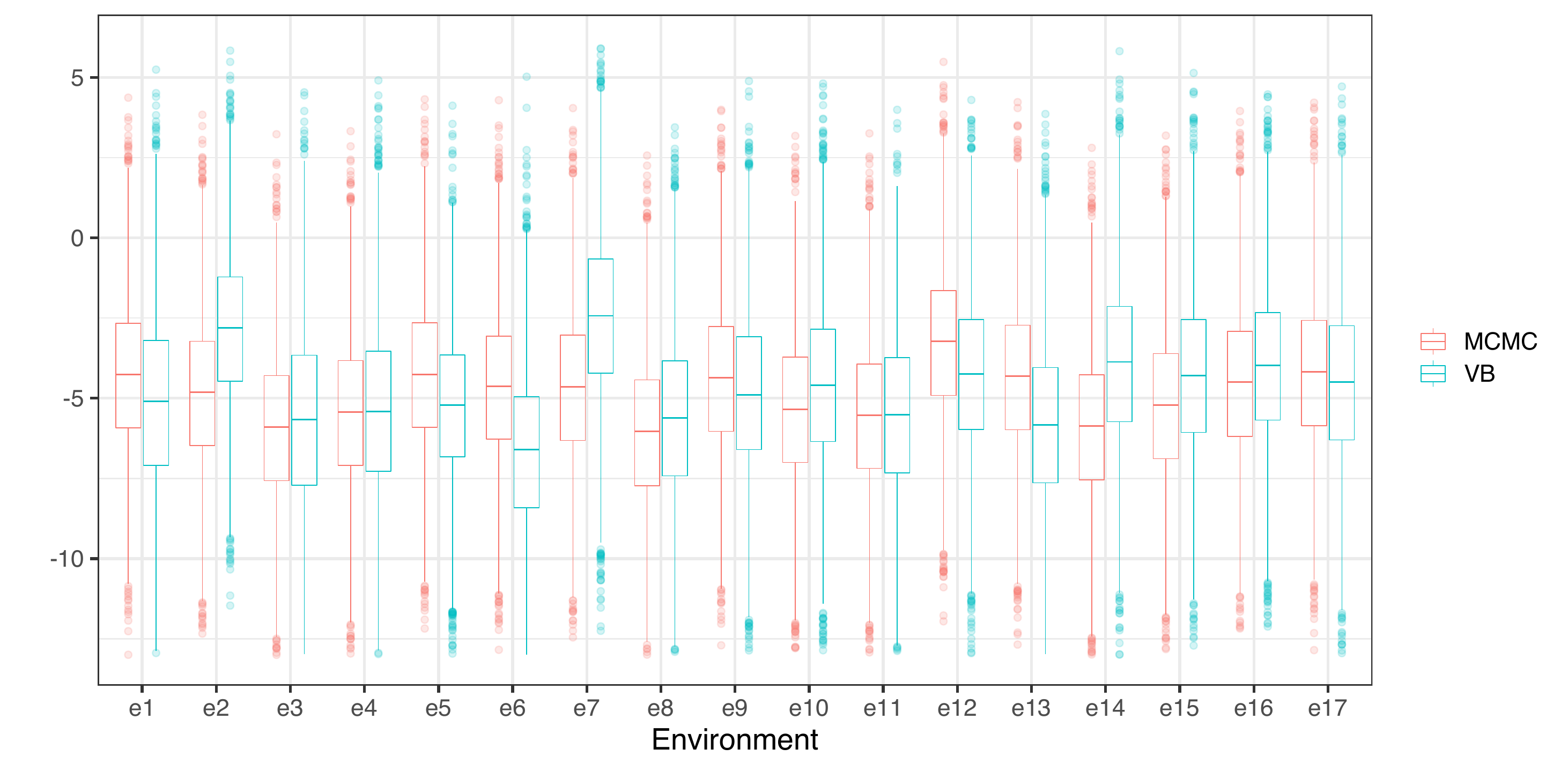}
			\caption{Posterior distributions of the 17 environments obtained via VI and MCMC (Q=1).}
        \label{fig:compmcmcvbenv}
\end{figure}

\section{Variational Updates}\label{VU}

\begin{enumerate}
    \item Variational distribution of $\mu$.
    \begin{eqnarray*}
    q(\mu) &\sim& \mathcal{N}(\mu_{q(\mu)}, \Sigma_{q(\mu)}^{-1});\\
    \mu_{q(\mu)} &=& \Sigma_{q(\mu)}^{-1}\left[\tilde{\tau}\left(\sumi\sumj y_{ij} - n_e\sumi\tilde{g_i} - n_g \sumj\tilde{e_j} - \sumi\sumj\sum_{q =1}^2 \tilde{\lambda}_q\tilde{\gamma}_{iq}\tilde{\delta}_{jq}\right) + \frac{\mu_\mu}{\sigma^2_\mu}\right]\\
    \Sigma_{q(\mu)} &=& n\tilde{\tau} + \frac{1}{\sigma^2_\mu}
    \end{eqnarray*}
    
    \item Variational distribution of $\bm{g}$.
       \begin{eqnarray*}
    q(\bm{g}) &\sim& \prodi \mathcal{N}(\mu_{q(g_i)}, \Sigma_{q(g_i)}^{-1});\\
    \mu_{q(g_i)} &=& \Sigma_{q(g_i)}^{-1}\left[\tilde{\tau}\left(\sumj y_{ij} - n_e\tilde{\mu} - \sumj\tilde{e_j} - \sumj\sum_{q =1}^2 \tilde{\lambda}_q\tilde{\gamma}_{iq}\tilde{\delta}_{jq}\right) \right]\\
    \Sigma_{q(g_i)} &=& n\tilde{\tau} + \frac{1}{\sigma^2_g}
    \end{eqnarray*}
    
    \item Variational distribution of $\bm{e}$.
           \begin{eqnarray*}
    q(\bm{g}) &\sim& \prodj \mathcal{N}(\mu_{q(e_j)}, \Sigma_{q(e_j)}^{-1});\\
    \mu_{q(e_j)} &=& \Sigma_{q(e_j)}^{-1}\left[\tilde{\tau}\left(\sumi y_{ij} - n_g\tilde{\mu} - \sumi\tilde{g_i} - \sumi\sum_{q =1}^2 \tilde{\lambda}_q\tilde{\gamma}_{iq}\tilde{\delta}_{jq}\right) \right]\\
    \Sigma_{q(e_j)} &=& n\tilde{\tau} + \frac{1}{\sigma^2_e}
    \end{eqnarray*}
    
    \item Variational distribution of $\lambda_1$.
     \begin{eqnarray*}
    q(\lambda_1) &\sim& \mathcal{TN}(\mu_{q(\lambda_1)}, \Sigma_{q(\lambda_1)}^{-1});\\
    \mu_{q(\lambda_1)} &=& \Sigma_{q(\lambda_1)}^{-1}\left[ \tau \gamma_{i1}\delta_{j1}(y_{ij}+\mu + g_i+e_j) + \lambda_2\gamma_{i1}\delta_{j1}\gamma_{i2}\delta_{j2}\right]\\
    \Sigma_{q(\lambda_1)} &=& \frac{\tilde{\tau}}{2}\sumij(\gamma_{i1^2}\delta_{j1}^2) + \frac{\lambda_1^2}{\sigma^2_\lambda }
    \end{eqnarray*}
    
    \item Variational distribution of $\lambda_2$.
    
    \begin{eqnarray*}
    q(\lambda_2) &\sim& \mathcal{TN}(\mu_{q(\lambda_2)}, \Sigma_{q(\lambda_2)}^{-1});\\
    \mu_{q(\lambda_2)} &=& \Sigma_{q(\lambda_2)}^{-1}\left[ \tau \gamma_{i2}\delta_{j2}(y_{ij}+\mu + g_i+e_j) + \lambda_1\gamma_{i1}\delta_{j1}\gamma_{i2}\delta_{j2}\right]\\
    \Sigma_{q(\lambda_2)} &=& \frac{\tilde{\tau}}{2}\sumij(\gamma_{i2^2}\delta_{j2}^2) + \frac{\lambda_2^2}{\sigma^2_\lambda}
    \end{eqnarray*}
\end{enumerate}
\end{appendix}

\end{document}